# Knowledge Sources for Word Sense Disambiguation

Eneko Agirre[1] and David Martinez[1]

[1] IxA NLP group, Basque Country University, 649 pk.,
E-20080 Donostia, Spain
{eneko, jibmaird}@si.ehu.es

**Abstract.** Two kinds of systems have been defined during the long history of WSD: principled systems that define which knowledge types are useful for WSD, and robust systems that use the information sources at hand, such as, dictionaries, light-weight ontologies or hand-tagged corpora. This paper tries to systematize the relation between desired knowledge types and actual information sources. We also compare the results for a wide range of algorithms that have been evaluated on a common test setting in our research group. We hope that this analysis will help change the shift from systems based on information sources to systems based on knowledge sources. This study might also shed some light on semi-automatic acquisition of desired knowledge types from existing resources.

## 1  Introduction

Research in Word Sense Disambiguation (WSD) has a long history, as long as Machine Translation. A vast range of approaches has been pursued, but none has been successful enough in real-world applications. The last wave of systems using machine learning techniques on hand-tagged corpora seems to have reached its highest point, far from the expectations raised in the past. The time has come to meditate on the breach between *principled* systems that have deep and rich hand-built knowledge (usually a Lexical Knowledge Base, LKB), and *robust* systems that use either superficial or isolated information.

Principled systems attempt to describe the desired kinds of knowledge and proper methods to combine them. In contrast, robust systems tend to use whatever lexical resource they have at hand, either Machine Readable Dictionaries (MRD) or light-weight ontologies. An alternative approach consists on hand-tagging word occurrences in corpora and training machine learning methods on them. Moreover, systems that use corpora without the need of hand-tagging have also been proposed. In any case, little effort has been made to systematize and analyze what kinds of knowledge have been put into play. We say that robust systems use *information sources*, and principled systems use *knowledge types*.

Another issue is the performance that one can expect from each information source or knowledge type used. Little comparison has been made, specially for knowledge types, as each research team tends to evaluate its system on a different experimental setting. The SENSEVAL competition [1] could be used to rank the knowledge types separately, but unfortunately, the systems tend to combine a variety of heuristics without separate evaluation. We tried to evaluate the contribution of each information source and knowledge type separately, testing each system in a common setting: the English sense inventory from WordNet 1.6 [2], and a test set comprising either all occurrences of 8 nouns in Semcor [3] or all nouns occurring in a set of 4 random files from Semcor.

This paper is a first attempt to systematize the relation between desired knowledge types and actual information sources, and to provide an empirical comparison of results on a common experimental setting. In particular, a broad range of systems that the authors have implemented is analyzed in context.

Although this paper should review a more comprehensive list of references, space requirements allow just a few relevant references. For the same reason, the algorithms are just sketched (the interested reader has always a pointer to a published reference), and the results are given as averages.

The structure of the paper is as follows. Section 2 reviews traditional knowledge types useful for WSD. Section 3 introduces an analysis of the information sources for a number of actual systems. Section 4 presents the algorithms implemented, together with the results obtained. Section 5 presents a discussion of the results, including future research directions. Finally, section 6 draws some conclusions.

## 2   Knowledge Types Useful for WSD

We classify the knowledge types useful for disambiguating an occurrence of a word based on Hirst [4], McRoy [5] and our own contributions. The list is numbered for future reference.

1. **Part of speech** (POS) is used to organize the word senses. For instance, in WordNet 1.6 *handle* has 5 senses as a verb, only one as a noun.
2. **Morphology**, specially the relation between derived words and their roots. For instance, the noun *agreement* has 6 senses, its verbal root *agree* 7, but not all combinations hold.
3. **Collocations**. The 9-way ambiguous noun *match* has only one possible sense in "*football match*".
4. **Semantic word associations**, which van be further classified as follows:
   a  **Taxonomical organization**, e.g. the association between *chair* and *furniture*.
   b  **Situation**, such as the association between *chair* and *waiter*.
   c  **Topic**, as between *bat* and *baseball*.
   d  **Argument-head relation**, e.g. *dog* and *bite* in "*the dog bite the postman*".
   These associations, if given as a sense-to-word relation, are strong indicators for a sense. For instance, in "*The chair and the table were missing*" the shared class in the taxonomy with *table* can be used to choose the furniture sense of *chair*.
5. **Syntactic cues**. Subcategorization information is also useful, e.g. *eat* in the "*take a meal*" sense is intransitive, but it is transitive in other senses.
6. **Semantic roles**. In "*The bad new will eat him*" the object of eat fills the *experiencer* role, and this fact can be used to better constrain the possible senses for eat.

7. **Selectional preferences**. For instance, *eat* in the "*take a meal*" sense prefers humans as subjects. This knowledge type is similar to the argument-head relation (4d), but selectional preferences are given in terms of semantic classes, rather that plain words.
8. **Domain**. For example, in the domain of sports, the "*tennis racket*" sense of *racket* is preferred.
9. **Frequency of senses**. Out of the 4 senses of *people* the general sense accounts for 90% of the occurrences in Semcor.
10. **Pragmatics**. In some cases, full-fledged reasoning has to come into play to disambiguate *head* as a *nail-head* in the now classical utterance "*Nadia swing the hammer at the nail, and the head flew off*" [4].

Some of these types are out of the scope of this paper: POS tagging is usually performed in an independent process and derivational morphology would be useful only for disambiguating roots.

Traditionally, the lexical knowledge bases (LKBs) containing the desired knowledge have been built basically by hand. McRoy, for instance, organizes the knowledge related to 10,000 lemmas in four inter-related components:

1. lexicon: core lexicon (capturing knowledge types 1, 2, 5 and 9) and dynamic lexicons (knowledge type 8)
2. concept hierarchy (including 4a, 6 and 7)
3. collocational patterns (3)
4. clusters of related definitions (sets of clusters for 4b and 4c)

Manual construction of deep and rich semantic LKBs is a titanic task, with many shortcomings. It would be interesting to build the knowledge needed by McRoy's system by semi-automatic means. From this perspective, the systematization presented in this paper can be also understood as a planning step towards the semi-automatic acquisition of such semantic lexicons.

## 3  Information Sources Used in Actual Systems

WSD systems can be characterized by the information source they use in their algorithms, namely MRDs, light-weight ontologies, corpora, or a combination of them [6]. This section reviews some of the major contributors to WSD (including our implementations). The following section presents in more detail the algorithms that we implemented and tested on the common setting. The systems are organized according to the major information source used, making reference to the knowledge types involved.

**MRDs** (4, 5, 7, 9). The first sense in dictionaries can be used as an indication of the most used sense (9). Other systems [7] [8] try to model semantic word associations (4) processing the text in the definitions in a variety of ways. Besides, [7] uses the additional information present in the machine-readable version of the LDOCE dictionary, subject codes (4a), subcategorization information (5) and basic selectional preferences (7). Unfortunately, other MRDs lack this latter kind of information.

**Ontologies** (4a). Excluding a few systems using proprietary ontologies, most systems have WordNet [2] as the basic ontology. Synonymy and taxonomy in WordNet provide the taxonomical organization (4a) used in semantic relatedness measures [9] [10].

**Corpora** (3, 4b, 4c, 4d, 5). Hand tagged corpora has been used to train machine learning algorithms. The training data is processed to extract features, that is, cues in the context of the occurrence that could lead to disambiguate the word correctly. For instance, Yarowsky [11] showed how collocations (3) could be captured using bigrams and argument-head relations. In the literature, easily extracted features are preferred, avoiding high levels of linguistic processing [11] [12] [13] [14]. In general, two sets of features are distinguished:

1. Local features, which use local dependencies (adjacency, small window, and limited forms of argument-head relations) around the target sense. The values of the features can be word forms, lemmas or POS. This set of features tries to use the following knowledge types without recognizing them explicitly: collocations (3), argument-head relations (4d) and a limited form of syntactic cues (5), such as adjacent POS.
2. Global features consist on bags of lemmas in a large window (50, 100 words wide) around the target word senses. Words that co-occur frequently with the sense would indicate that there is a semantically association of some kind (usually related to the situation or topic, 4b, 4c).

During testing, a machine learning algorithm is used to compare the features extracted from the training data to the actual features in the occurrence to be disambiguated. The sense with the best matching features is selected accordingly.

**MRD and ontology combinations** (4a, 4b, 4c) have been used to compensate for the lack of semantic associations in existing ontologies like WordNet. For instance, [15] combines the use of taxonomies and the definitions in WordNet yielding a similarity measure for nominal and verbal concepts which are otherwise unrelated in WordNet. The taxonomy provides knowledge type 4a, and the definitions implicitly provide 4b and 4c.

**MRD and corpora combinations** (3, 4b, 4c, 4d, 5). [16] uses the hierarchical organization in Roget's thesaurus to automatically produce sets of salient words for each semantic class. These salient words are similar to McRoy's clusters [5], and could capture both situation and topic clusters (4b, 4c). In [17], seed words from a MRD are used to bootstrap a training set without the need of hand-tagging (all knowledge types used for corpora could be applied here, 3, 4b, 4c, 4d, 5).

**Ontology and corpora combinations** (3, 4b, 4c, 4d, 5, 7). In an exception to the general rule, selectional preferences (7) have been semi-automatically extracted and explicitly applied to WSD [9] [18]. The automatic extraction involved the combination of parsed corpora to construct sets of e.g. nouns that are subjects of an specific verb, and a similarity measure based on a taxonomy is used to generalize the sets of nouns to semantic classes. In a different approach [14], the information in WordNet has been used to build automatically a training corpus from the web (thus involving knowledge types 3, 4b, 4c, 4d, 5). A similar technique has been used to build topic signatures, which try to give lists of words topically associated for each concept [19].

For some of the knowledge types we could not find implemented systems. We are not aware of any system using semantic roles, or pragmatics, and domain information is seldom used.

## 4   Experimental Setting and Implementation of Main Algorithms

A variety of the WSD algorithms using the information sources mentioned in the previous section have been implemented in our research team. These algorithms are presented below, with a summary in Table 1, but first, the experimental setting will be introduced.

The algorithms were tested in heterogeneous settings: different languages, sense inventories, POS considered, training and test sets. For instance, a set of MRD-based algorithms where used to disambiguate salient words in dictionary definitions for all POS in a Basque dictionary. Nevertheless, we tried to keep a common experimental setting during the years: The English sense inventory taken from WordNet 1.6 [2], and a test set comprising either all occurrences of 8 nouns (*account, age, church, duty, head, interest, member* and *people*) in Semcor [3] or all polysemous nouns occurring in a set of 4 random files from Semcor (*br-a01, br-b20, br-j09* and *br-r05*). Some algorithms have been tried on the set of words, others on all words in the 4 files, others on both. Two algorithms have been tested on Semcor 1.4, but the results are roughly similar (Table 1 shows the random baselines for both versions of WordNet).

It has to be noted that WordNet plays both the roles of the ontology (e.g. offering taxonomical structure) and the MRD (e.g. giving textual definitions for concepts).

### 4.1   Algorithms Based on MRD

In [8] we present a set of heuristics that can be used alone or in combination. Basically, these heuristics use the definitions in a Spanish and a French MRD in order to sense-disambiguate the genus terms in the definitions themselves. Some of the simplest techniques have been more recently tried on the common test set and can thus be compared. These heuristics are the following:

**Main sense.** The first sense of the dictionaries is usually the most salient. This fact can be used to approximate the most frequent sense (MFS, 9). In our implementation, the word senses in WordNet are ordered according to frequency in Semcor, and thus the first sense corresponds to the MFS. The figures are taken from [14] [19]. Table 1 shows the results for both the 8-noun and 4-file settings. The MFS can be viewed as the simplest learning technique, and it constitutes a lower bound for algorithms that use hand-tagged corpora.

**Table 1.** Summary of knowledge sources and results of algorithms implemented. The first colums shows the information source, the second the specific information used and the related knowledge types and the third the algorithm used. Finally evaluation is given for the two test sets using precision (correct answers over all examples) and coverage (all answers over all examples). Note: *global content here includes the combination of local and global context.

| Information source | Knowledge types | Algorithm | Results (prec. / cov.) | | | |
|---|---|---|---|---|---|---|
| | | | 8 nouns | | 4 files | |
| Random baseline | - | WordNet 1.4 | - | - | .30 | 1.0 |
| | | WordNet 1.6 | .19 | 1.0 | .28 | 1.0 |
| MRD | Main sense 9 | | .69 | 1.0 | .66 | 1.0 |
| | Definition 4 | Overlap | .42 | 1.0 | - | - |
| Ontology | Hierarchy 4a | Conc. Density | - | - | .43 | .80 |
| Corpora | Most freq. sense 9 | | .69 | 1.0 | .66 | 1.0 |
| | Local context 3 4d 5 | Decision lists | .78 | .96 | * | * |
| | Syntactic cues 5 | " | .70 | .92 | - | - |
| | Arg.-head relations 4d | " | .78 | .69 | - | - |
| | Global context 4b 4c | " | .81 | .87 | .69* | .94* |
| MRD + Corpora | Semantic classes 4b 4c | Mutual info. | - | - | .41 | 1.0 |
| Ontology + Corpora | Selectional pref. 7 | Probability | .63 | .33 | .65 | .31 |
| | Topic signatures 4b 4c | $Chi^2$ | .26 | .99 | - | - |
| | Aut. tagged corp. 3 4 5 | Decision lists | .13 | .71 | - | - |

**Definition overlap.** In the most simple form the overlap between the definitions for the word senses of the target word and the words in the surrounding context is used [19]. This is a very limited form of knowledge type 4, but its precision (cf. Table 1 for results on the 8-noun set) is nevertheless halfway between the random baseline and the MFS. We have also implemented more sophisticated ways of using the definitions, as co-occurrences, co-occurrence vectors and semantic vectors [8], but the results are not available for the common experimental setting.

### 4.2 Algorithms Based on Ontologies

Conceptual density [10] [20] is a measure of concept-relatedness based on taxonomies that formalizes the common semantic class (knowledge type 4a). The implementation for WordNet was tested on the 4-file test set using WordNet version 1.4.

### 4.3 Algorithms Based on Corpora

Hand tagged corpora has been used to train machine learning algorithms. We are particularly interested in the features used, that is, the different knowledge sources used by each system [11] [12] [13] [14]. We have tested a comprehensive set of features [14] [21] which for the sake of this paper we organized as follows:

- Local context features comprise bigrams and trigrams of POS, lemmas and word forms, as well as a bag of the words and lemmas in a small window comprising 4 words around the target [14]. These simple features involve knowledge about collocations (3), argument-head relations (4d) and limited syntactic cues (5).
- In addition to the basic feature set, syntactic dependencies were extracted to try to model better syntactic cues (5) and argument-head relations (4d) [21]. The results for both are given separately in Table 1.
- The only global feature in this experiment is a bag of word for the words in the sentence (knowledge types 4b, 4c) [14].

We chose to use one of the simplest yet effective way to combine the features: decision lists [22]. The decision list orders the features according to their log-likelihood, and the first feature that is applicable to the test occurrence yields the chosen sense. In order to use all the available data, we used 10-fold cross-validation. Table 1 shows the results in the 8-noun set for each of the feature types. In the case of the 4-file set, only the combined result of local and global features is given.

### 4.4 Algorithms Based on a Combination of MRD and Corpora

In the literature, there is a variety of ways to combine the knowledge in MRDs with corpora. We implemented a system that combined broad semantic classes with corpora [16] [20], and disambiguated at a coarse-grained level (implicitly covering knowledge types 4b and 4c). It was trained using the semantic files in WordNet 1.4 and tested on the 4-file setting only. In order to compare the performance of this algorithm that returns coarse-grained senses with the rest, we have estimated the fine-grained precision choosing one of the applicable fine-grained senses at random. The results are worse than reported in [16], but it has to be noted that the organization in lexical files is very different from the one in Roget's.

### 4.5 Algorithms based on ontologies and corpora

Three different approaches have been tried in this section:

- **Selectional preferences** (7). We tested a formalization that learns selectional preferences for classes of verbs [18] on subject and object relations extracted from Semcor. In this particular case, we used the sense tags in Semcor, partly to compensate for the lack of data. The results are available for both test settings. Note that the coverage of this algorithm is rather low, due to the fact that only 33% of the nouns in the test sets were subjects or objects of a verb.
- **Learning topic signatures from the web** (4b, 4c). The information given in WordNet for each concept is used to feed a search engine and to retrieve a set of training examples for each word sense. These examples are used to induce a set of words that are related to a given word sense in contrast with the other senses of the target word [19]. Topic signatures are constructed using the most salient words as given by the Chi$^2$ measure. Table 1 shows the results for the 8-noun setting, which are slightly above the baseline.
- **Inducing a training corpus from the web** (3, 4b, 4c, 4d, 5). In a similar approach, the training examples retrieved from the web are directly used to train decision lists, which were tested on the 8 noun set [14]. The results are very low on average, but the variance is very high, with one word failing for all test samples, and others doing just fine.

## 5   Discussion and Future Directions

From the comparison of the results, it is clear that algorithms based on hand-tagged corpora provide the best results. This is true for all features (local, syntactic cues, argument-head relations, global), including the combination of hand-tagged corpora with taxonomical knowledge (selectional restrictions). Other resources provide more modest results: conceptual density on ontologies, definition overlap on MRDs, or the combination of MRD and corpora. The combinations of corpora and ontologies that try to acquire training data automatically are promising, but current results are poor.

If the results are analyzed from the perspective of knowledge types, we can observe the following:

a. Collocations (3) are strong indicators if learned from hand-tagged corpora.
b. Taxonomical information is very weak (4a).
c. Semantic word associations around topic (4b) and situation (4c) are powerful when learned from hand-tagged corpora (but difficult to separate one from the other). Associations learned from MRDs can also be useful.
d. Syntactic cues (5) are reliable when learned from hand-tagged corpora.
e. The same applies for selectional preferences (7), but in this case the applicability, is quite low. It is matter of current experimentation to check whether the results are maintained when learning from raw corpora.
f. MFS (9) is also a strong indicator that depends on hand-tagged data.
g. POS (1), morphology (2), semantic roles (6), domain (8) and pragmatic (10) knowledge types have been left aside.

The results seem to confirm McRoy's observation that collocations and semantic word associations are the most important knowledge types for WSD [5], but we have noticed that syntactic cues are equally reliable. Moreover, the low applicability of selectional restrictions was already noted by McRoy [5].

All in all, hand-tagged corpora seems to be the best source for the automatic acquisition of all knowledge types considered, that is, collocations (3), semantic associations (situation 4b, topic 4c and argument-head relation 4d), syntactic cues (5), selectional restrictions (7) and MFS (9). Only taxonomic knowledge (4a) is taken from ontologies. In some cases it is difficult to interpret the meaning of the features extracted form corpora, e.g. whether a local feature reflects a collocation or not, or whether a global feature captures a topical association or not. This paper shows some steps to classify the features according to the knowledge type they represent.

However strong, hand tagged algorithms seem to have reached their maximum point, far from the 90% precision. These algorithms depend on the availability of hand-tagged corpora, and the effort to hand-tag the occurrences of all polysemous words can be a very expensive task, maybe comparable to the effort needed to build a comprehensive LKB. Semcor is a small sized corpus (around 250,000 words), and provides a limited amount of training data for a wide range of words. This could be the reason for the low performance (69% precision) when tested on the polysemous nouns in the 4 Semcor files. Unfortunately, training on more examples does not always raise precision much: in experiments on the same sense inventory but using more data, the performance raised from 73% precision for a subset of 5 nouns to only 75%. In the first SENSEVAL [1] the best systems were just below the 80% precision.

We think that future research directions should resort to all available information sources, extending the set of features to more informed features. Organizing the information sources around knowledge types would allow for more powerful combinations. Another promising area is that of using bootstrapping methods to alleviate or entirely eliminate the need of hand-tagging corpora. Having a large number of knowledge types at hand can be the key to success in this process.

The work presented in this paper could be extended, specially to cover more information sources and algorithms. Besides, the experimental setting should include other POS apart from nouns, and all algorithms should be tested on both experimental settings. Finally, it could be interesting to do a similar study that focuses on the different algorithms using principled knowledge and/or information sources.

## 6  Conclusions

We have presented a first attempt to systematize the relation between principled knowledge types and actual information sources. The former provide guidelines to construct LKBs for theoretically motivated WSD systems. The latter refer to robust WSD systems, which usually make use of the resources at hand: MRD, light-weight ontologies or corpora. In addition, the performance of a wide variety of knowledge types and algorithms on a common test set has been compared.

This study can help to understand which knowledge types are useful for WSD, and why some WSD systems perform better than others. We hope that in the near future, research will shift from systems based on information sources to systems based on knowledge sources. We also try to shed some light on the possibilities for semi-automatic enrichment of LKBs with the desired knowledge from existing resources.

## Acknowledgements

Some of the algorithms have been jointly developed in cooperation with German Rigau. David Martinez has a scholarship from the Basque Country University. This work was partially funded by FEDER funds from the European Commission (*Hiztegia 2001* project) and by MCYT (*Hermes* project).